\pdfoutput=1

\documentclass[11pt]{article}

\usepackage{EMNLP2023}

\usepackage{times}
\usepackage{latexsym}
\usepackage{float}

\usepackage[T1]{fontenc}

\usepackage[utf8]{inputenc}

\usepackage{microtype}

\usepackage{inconsolata}
\usepackage{graphicx}
\usepackage{enumitem}
\usepackage{tikz}
\usepackage{amsmath}
\usetikzlibrary{3d}

\setlist{leftmargin=2.5mm}

\usepackage{titlesec}
\titlespacing*{\subsection}{0pt}{0.15\baselineskip}{0.05\baselineskip}
\titlespacing*{\section}{0pt}{0.6\baselineskip}{0.5\baselineskip}

\usepackage{balance}

\title{Accurate and Data-Efficient Toxicity Prediction when Annotators Disagree}

\author{Harbani Jaggi*\\UC Berkeley \And Kashyap Murali*\\UC Berkeley \And Eve Fleisig\\UC Berkeley \And Erdem Bıyık\\USC}

\def\embedding\space{embedding-based architecture} 

\begin{document}
\maketitle
\begin{abstract}

When annotators disagree, predicting the labels given by individual annotators can capture nuances overlooked by traditional label aggregation. We introduce three approaches to predicting individual annotator ratings on the toxicity of text by incorporating individual annotator-specific information: a neural collaborative filtering (NCF) approach, an in-context learning (ICL) approach, and an intermediate \embedding\space. We also study the utility of demographic information for rating prediction. NCF showed limited utility; however, integrating annotator history, demographics, and survey information permits both the \embedding\space\ and ICL to substantially improve prediction accuracy, with the \embedding\space\; outperforming the other methods. We also find that, if demographics are predicted from survey information, using these imputed demographics as features performs comparably to using true demographic data. This suggests that demographics may not provide substantial information for modeling ratings beyond what is captured in survey responses. Our findings raise considerations about the relative utility of different types of annotator information and provide new approaches for modeling annotators in subjective NLP tasks.


\end{abstract}


\section{Introduction}

Disagreement among data annotators can reveal nuances in NLP tasks that lack a simple ground truth, such as hate speech detection. For instance, what one group of annotators deems acceptable might be considered offensive by another. The current standard for resolving such disagreement, aggregation via majority voting, casts aside variance in annotator labels as noise, when in subjective tasks this variance is key to understanding the perspectives that arise from the annotators’ individuality and backgrounds.

To address this problem, recent research has explored alternatives to majority voting. Most notably, studies have taken the approach of predicting the ratings of individual annotators  \cite{davani2022dealing, fleisig2023majority, gordon-etal-2022-jury}. We aim to improve the prediction of rating behavior, guided by the following questions:

\begin{itemize}[nosep]
    \item \textbf{Does incorporating annotator information via collaborative filtering,  \embedding\space, or in-context learning improve downstream rating predictions?}
    \item \textbf{What annotator information best informs toxicity rating predictions? Do demographics provide useful information beyond what survey information can provide?}
\end{itemize}

We proposed and tested a neural collaborative filtering (NCF) module, an \embedding\space, and an in-context learning (ICL) module for individual rating prediction. First, we incorporated NCF to the classification head of a RoBERTa-based model \cite{liu2019roberta}. Embedded annotator information\footnote{The annotator information used is a combination of demographic information, survey information, and annotator rating history.} was combined with a separate embedding of annotators' rating history to predict individual annotator toxicity ratings. Secondly, we used embedding models to encode annotator information, which was then used to predict toxicity ratings. Lastly, we prompted LLMs such as Mistral \citep{jiang2023mistral} and GPT-3.5 \citep{brown2020language} to study different ways of integrating annotator information.

Our findings indicate that while NCF does not outperform baseline models, ICL and our \embedding\space\; improve performance, with the \embedding\space\; significantly outperforming all other approaches tested. In addition, our research on the effectiveness of demographic information as a feature indicates that imputing demographics from survey data performs similarly to using direct demographic inputs, suggesting that survey responses already capture the relevant demographic information for rating prediction. This suggests that, on this task, demographics have little predictive power beyond what survey information provides.


\section{Motivation and Related Work}

Our work is fundamentally motivated by the need for alternatives to majority-vote label aggregation in NLP tasks. \citet{pavlick2019inherent} find that disagreement among annotators is partially attributed to differences in human judgment. \citet{basile2021we} underscore the importance of the consideration of a system’s output over instances where annotators disagree.

Newer work in this field aims to directly model individual annotator rating behavior. \citet{davani2022dealing} employ a multi-task based approach, where predicting each annotators’ judgment is a subtask to their larger architecture. \citet{fleisig2023majority} use a RoBERTa-based model to predict an individual annotators’ ratings. \citet{gordon-etal-2022-jury} put together a jury of annotators, predicting individual judgments.


For the individual annotator rating prediction task, \citet{deng2023you} create individual annotator embeddings and annotation embeddings. This idea of learning embeddings based on user-specific data has been applied in various domains successfully, e.g., imitation learning \citep{beliaev2022imitation} or recommendation systems \citep{biyik2023preference}.

Collaborative filtering (CF) learns user embeddings based on their past behaviors \citep{bokde2015matrix}. \citet{he2017neural} show that neural collaborative filtering (NCF) offers better performance than more naive CF implementations. This motivates our NCF approach to learning annotator embeddings. Intuitively, this approach would be effective in learning deeply rooted preferences and behaviors of annotators. Thus, we hypothesized that this method would more accurately predict individual annotator ratings.

Several recent approaches use sociodemographic traits of individual annotators to learn for the rating prediction task \citep{fleisig2023majority, davani2022dealing}, but \citet{andrus2021we} warn that legal and organizational constraints, such as privacy laws and concerns around self-reporting, often make collecting demographic data challenging. \citet{gupta2018proxy} suggest using semantically related features in the absence of sensitive demographic data. For instance, in the absence of gender information, \citep{zhao2019conditional} use other demographic features -- age, relation, and marital status -- for their prediction task. This work motivates our objective of incorporating auxiliary annotator information (survey information and annotator history) in the prediction task.

Lastly, \citet{orlikowski-etal-2023-ecological} challenge the utility of demographic information, since they do not find strong evidence that explicitly modeling demographics helps to predict annotation behavior. In concurrent work, \citet{hu2024quantifying} argue that there is an inherent limit to how much predictive power can be provided by demographics. Their findings indicate that while incorporating demographic variables can provide modest improvements in prediction accuracy, these gains are often constrained by the relatively low variance explained by these variables. This motivates our final objective, studying the efficacy of demographics as a useful mediating variable for rating prediction.

\section{Approach}

Our approach includes creating three separate modules based on neural collaborative filtering (NCF), an \embedding\space, and in-context learning (ICL). We evaluate each approach's efficacy in predicting annotator rating behavior. The latter two modules are used to investigate our second research question; we integrate different ablations of annotator information as input to the rating prediction models to study their effect on toxicity rating prediction.

We used \citet{KUMAR2021DESIG}'s dataset to evaluate the performance of our rating prediction modules. This dataset consists of sentences rated for toxicity (0 = least toxic, 4 = most toxic). Each sentence has been labeled by 5 annotators and each annotator has labeled 20 distinct sentences. For each annotator, the dataset contains their rating behavior; demographic information (race, gender, importance of religion, LGBT status, education, parental status, and political stance); and survey information, e.g., their preferred forums, social media, whether they have seen toxic content, if they think toxic content is a problem, and their opinion on whether technology impacts peoples' lives.

For ablations, we took distinct combinations of annotator information (rating history, demographics, survey information) along with the text to be rated, assessing the impact of each on the model's performance. To study whether demographics are a necessary feature for predicting annotator ratings, we also used a separate model to predict annotator demographics using rating history and survey information and applied these predicted demographics as input for our ablations.

For all three methods, we used Mean Absolute Error (MAE) of predicting individual annotators' ratings as the evaluation metric, allowing us to quantify the performance of different model configurations.


\subsection{Neural Collaborative Filtering}

\begin{figure}
    \centering
    \includegraphics[width=0.95\linewidth]{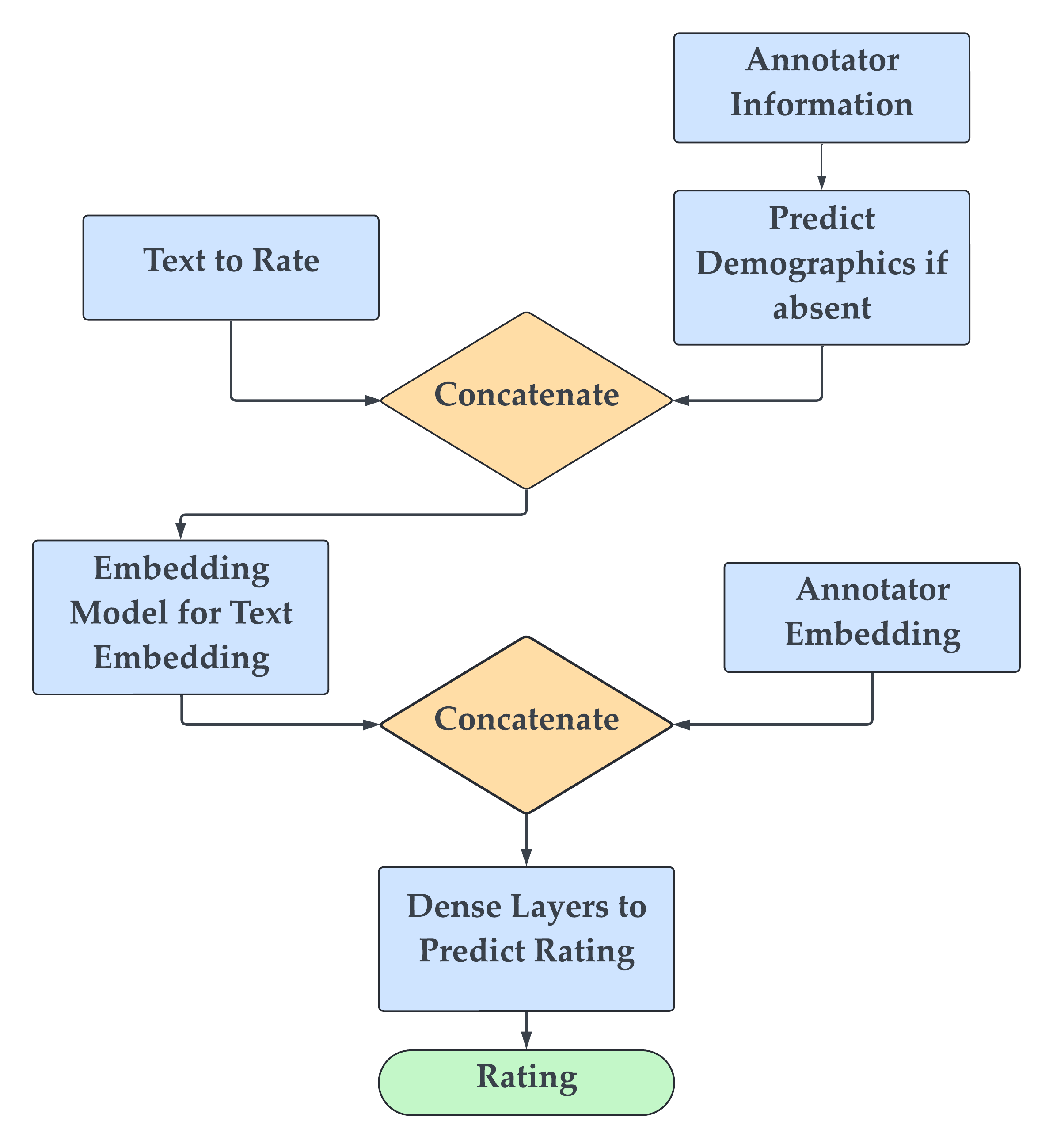}
    \caption{Design of our neural collaborative filtering (NCF) architecture. Annotator information and the text being rated were passed into an embedding model, then concatenated with the annotator embedding, and passed through a series of dense layers to predict the  rating.}
    \label{fig:ncf-arch}
\end{figure}

Our NCF method integrates textual and annotator-specific information to predict annotator ratings for the toxicity detection task (Figure \ref{fig:ncf-arch}). We aimed to create both a textual embedding and an annotator embedding for each (text, annotator) pair and capture latent interactions between both entities by using a hybrid neural architecture inspired by neural collaborative filtering. The goal was to learn more complex, non-linear relationships between annotator preferences and the text itself to more accurately predict an annotator's toxicity rating.

To create embedded representations of the textual information which has ranging levels of toxicity, we leveraged a RoBERTa model \citep{liu2019roberta}  fine-tuned on the Jigsaw Toxic Comment Classification Challenge dataset \citep{jigsaw-toxic-comment-classification-challenge} and the hate speech detection datasets introduced by \citet{KUMAR2021DESIG}. In parallel, we initialized and stored random embeddings for each annotator in the RoBERTa classification head. During training, these embeddings were concatenated with text embeddings and passed through 4 dense layers before predicting the rating. 

In developing this hybrid model architecture, we explored variations in the dimensionality of the annotator embeddings, methods for integrating the sentence and annotator embeddings, and the impact of freezing the RoBERTa model (Appendix \ref{sec:approaches_taken} describes variations tested).




\subsection{Embedding-Based Architecture}

\begin{figure}[h!]
    \centering
    \includegraphics[width=0.9\linewidth]{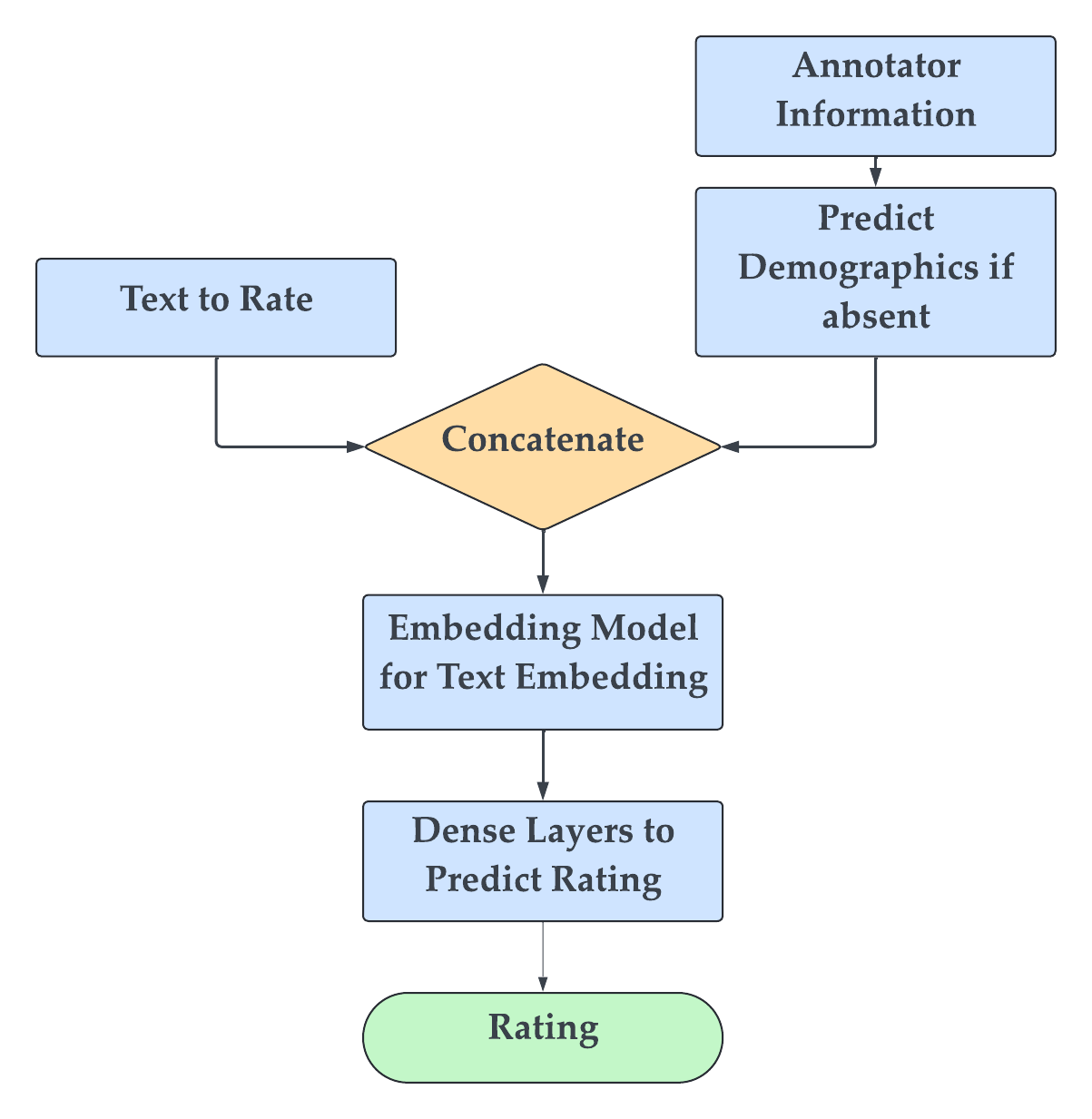}
    \caption{Design of our \embedding\space. }
    \label{fig:embedding-arch}
\end{figure}

We generated embeddings for the concatenated annotator information and the current text to be rated using two text embedding models, OpenAI's text-embedding-3-small and text-embedding-3-large. These embeddings then served as input for a custom model with multiple fully connected layers, which was trained to predict toxicity ratings based on the extracted features (Figure \ref{fig:embedding-arch}).

\begin{table*}[htbp]
\fontsize{9}{10}\selectfont
\centering
\begin{tabular}{p{5cm}ccp{3.3cm}p{3.3cm}}
\hline
\textbf{Model} & \textbf{Mistral} & \textbf{GPT 3.5} &  \textbf{text-embedding-3-small} & \textbf{text-embedding-3-large} \\
\hline
Text only & 0.78 & 0.81 & 0.76 & 0.75 \\
+ demo.& 0.76 & 0.79 & 0.73 & 0.71 \\
+ demo. + history& 0.75 & 0.78 & 0.73 & 0.69 \\
+ history& 0.73 & 0.75 & 0.70 & 0.66 \\
+ survey& 0.73 & 0.75 & 0.70 & 0.70 \\
+ demo. + survey& 0.71 & 0.73 & 0.68 & 0.64 \\
+ history + survey & 0.70 & 0.73 & 0.67 & 0.69 \ \\
+ predicted demo. + history + survey & 0.70 & 0.74 & 0.67 & 0.66 \\
+ demo. + history + survey& \textbf{0.69} &\textbf{ 0.72} & \textbf{0.66} & \textbf{0.61 }\\
\hline
\end{tabular}
\vspace{-7px}
\caption{\label{model-results-final}
Comparison of mean absolute error across different model configurations for the test set (with or without annotator demographics, rating history, and survey responses). Both ICL and \embedding\space s improve on the baseline, with \embedding\space s performing best.
}
\vspace{-7px}
\end{table*}

\subsection{In-Context Learning}

Our in-context learning architecture prompts a language model to process a range of combinations of annotator information. Each combination serves as input to the model (Mistral or GPT-3.5), enabling it to account for the specific context of the annotator when predicting toxicity ratings. The model was prompted to generate predictions based on the contextual information provided. This approach aims to enhance the model’s ability to make informed predictions by integrating diverse sources of information relevant to the rating task.  A sample prompt of this approach is shown in  Figure \ref{fig:toxicity_prompt}.



\section{Results}

Our three approaches predicted annotators’ toxicity ratings on a scale from 0 to 4, based on both textual data and various combinations of annotator-specific information (demographics, survey responses, rating history). We also examine how well these models handle predicted demographic data rather than using the ground truth demographic values for each annotator. This helps to assess the data efficiency and effect of demographics as an input to the rating prediction task.

For our ablations that studied the improvement on rating predictions, we compared our results to previous baselines that predicted ratings of annotators using the same dataset.

\textbf{Q1: Does incorporating annotator information via collaborative filtering, the \embedding\space, or in-context learning improve downstream rating predictions?}


Our \embedding\space\; outperformed all other experiments with an MAE of 0.61; the best ICL approach (with Mistral) reached an MAE of 0.69. Both the ICL approach and \embedding\space\; outperform the most recent baseline for the dataset \cite{fleisig2023majority} and the \embedding\space\; matches the best previous MAE on this dataset \citep{gordon-etal-2022-jury}. The best-performing models use all available annotator-specific information as input (annotator demographics, survey information, and historical rating data). At its best, our ICL configuration with Mistral had an MAE of 0.69 (using annotator demographics, survey information, and historical rating data). The NCF approach had consistently poorer results, with a best MAE of 0.79 when including all annotator-specific information.

When creating the NCF architecture, we tested several variations. We first created a baseline from which we compared different outputs of our NCF module. Evaluating the finetuned RoBERTa model with all annotator-specific information as input along with the text to be rated yielded a baseline MAE of 0.81. We experimented with integrating embeddings through dot product vs.~concatenation, freezing RoBERTa during the training process, and placing the collaborative filtering task in different parts of the RoBERTa architecture. Our best performing model froze the pretrained RoBERTa model, used concatenation, and placed the collaborative filtering piece in the classification head. However, it was only able to achieve an MAE of 0.80, not significantly improving on our baseline.

Our  \embedding\space\;consistently outperformed other approaches on every ablation, suggesting that a hybrid feature-extraction approach most effectively uses annotator-specific information in rating predictions.


\textbf{Q2: What annotator information best informs toxicity rating predictions? Do demographics provide useful information beyond what survey information can provide?}
\begin{figure}
    \centering
    \includegraphics[width=0.49\textwidth]{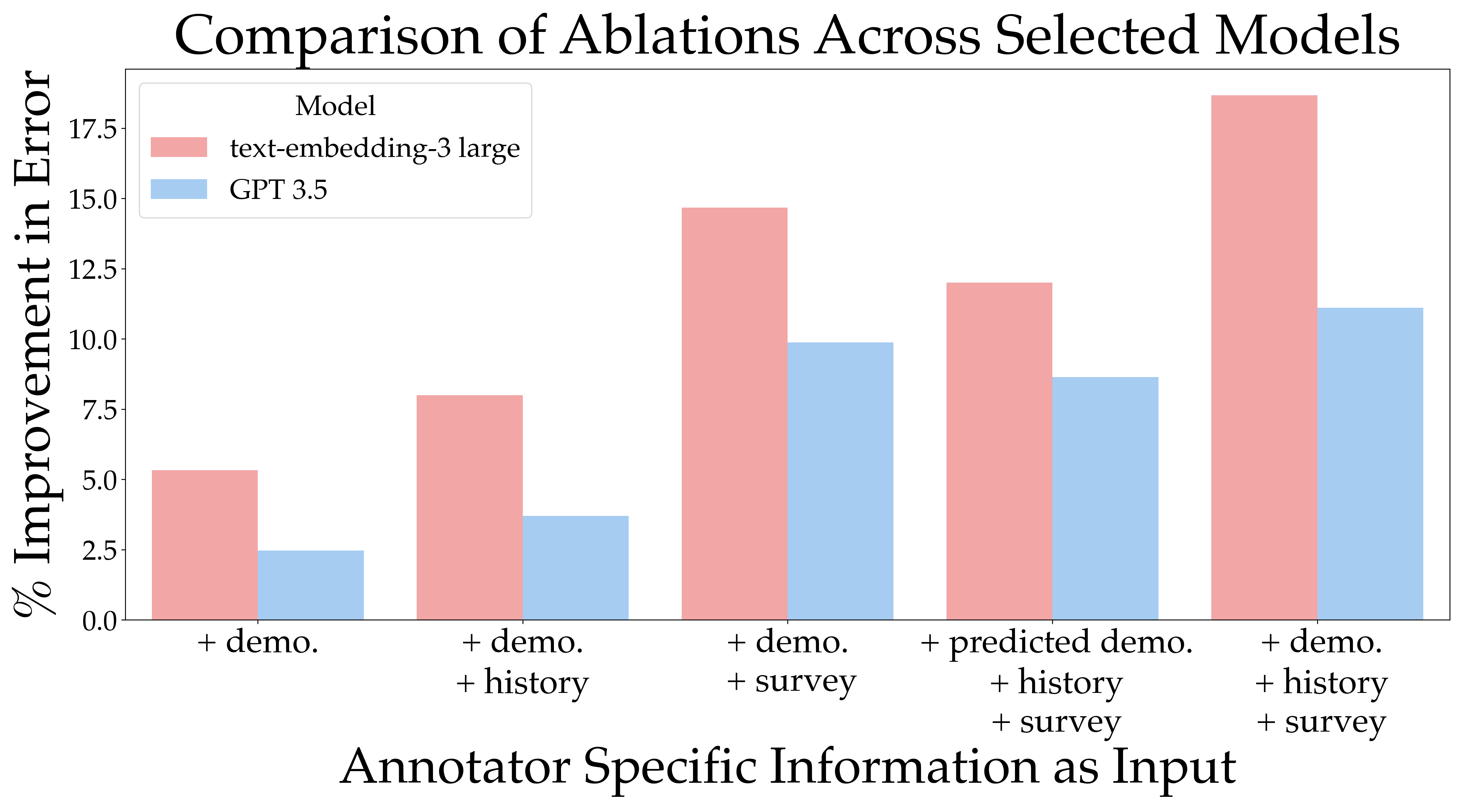}
    \caption{Comparison of  MAE improvement with varying amounts of annotator input across all models. The text-embedding-3-large model consistently outperforms all other models and has most improvement on its own baseline.}
    \label{fig:ablation_comparison}
\end{figure}

Incorporating demographic information improves performance over using only survey information, rating history, or both across ablations. However, we find that much of this gap can be compensated for by distilling demographic information out of survey information. 
Compared to the text-only baseline, incorporating predicted demographics with survey information and annotator history achieved MAE reductions of 10.26\% with Mistral, 8.64\% with GPT-3.5, 11.84\% with text-embedding-3-small, and 12\% with text-embedding-3-large. Replacing true demographic information with predicted demographic information results in nearly as strong performance for Mistral, GPT-3.5, and text-embedding-3-small.

Incorporating predicted demographics alongside survey information and annotator history notably improves accuracy. 
This occurs despite the fact that the accuracy of predicted demographics varies widely (highest for race and gender, but near-random for some demographics; see Table \ref{table:generated-data-results}). Although the true demographics are somewhat helpful, annotator ratings can be effectively predicted without direct demographic data. This finding suggests that detailed demographic data may not be especially useful as a feature in individual rating prediction, beyond what can be inferred from individual preferences in survey responses.

\textbf{Predicting Demographics.} The performance of predicting demographics was evaluated across various configurations (Table \ref{table:generated-data-results}). The baseline approach incorporating only survey information achieved the highest accuracies, with 47\% for race and 63\% for gender. Combining survey information with text slightly reduced the performance, potentially indicating the noise that the text to be rated added. The majority class approach is indicated as a baseline comparison to highlight the performance improvements for the different categories.

Our findings indicate that successively incorporating annotator demographics, rating history, and survey information improves performance for nearly all configurations tested (Table \ref{model-results-final}). Overall, the comprehensive model incorporating demographics, annotator history, and survey data consistently outperformed other configurations, demonstrating the value of integrating multiple data sources for demographic and rating predictions.

\section{Conclusion}

Leveraging the \embedding\space\;and ICL methods substantially improved toxicity rating predictions. NCF, by contrast, was not a competitive method for predicting ratings. Incorporating annotator information significantly enhances model performance. The best-performing \embedding\space\;achieved the lowest MAE of 0.61 by integrating demographics, annotator history, and survey data. This suggests that personalized predictions based on individual annotator preferences can lead to more accurate outcomes. Meanwhile, the ability to predict some demographics from survey information, and the fact that these imputed demographics nearly match performance with the true demographics, suggest that although demographics are helpful, individual annotator ratings can be predicted effectively without demographic data. This finding suggests that some  differences in annotator opinions may be best captured by modeling individual preferences rather than demographic trends. In addition, the effectiveness of our \embedding\space\;suggests that it could help to inform future frameworks for annotator rating prediction.


\section{Limitations}

While our study advances the accuracy of annotator rating predictions, several limitations exist. The generalizability of our findings is limited to English text from the U.S. and Canada, which hinders applicability in other linguistic and cultural contexts. The integration of detailed annotator information poses ethical and privacy risks and can amplify existing biases in the data. Additionally, the complexity and computational demands of our models challenge scalability and interpretability. Future research should address these issues to enhance the robustness and fairness of predictive models in subjective NLP tasks. It should also focus on expanding these methods to other domains and exploring the ethical implications of incorporating inferred data for predictions. By continuing to refine these approaches, we can develop more accurate and reliable models that better capture the complexities of human behavior and preferences.

\section{Ethical Considerations}
We found that individual ratings can be predicted well without demographic information. This is helpful in that it permits individualized rating prediction without collecting demographic information. Unfortunately, that does not mean the ratings are predicted \textit{independent} of demographic information: in fact, we also found that survey information is a close enough proxy that demographics can be predicted with substantially better than random accuracy, especially for race and gender, off of survey information responses. Incorporating these predicted demographics further improves accuracy. However, our finding thus uncovered the potential privacy issue that collecting seemingly innocuous survey information data carries the risk of revealing annotator demographics. This suggests that future research in this area must proceed with caution: collecting or inferring demographic information improves prediction accuracy, but risks tokenism (where opinions within a demographic group are assumed to be homogeneous). Instead, future research could identify survey information questions that help to improve rating prediction but do \textit{not} risk revealing annotator demographics.

\balance
\bibliography{anthology,custom}
\bibliographystyle{acl_natbib}

\clearpage
\appendix

\section{Appendix}
\label{sec:dev_results}
\begin{table*}
\fontsize{9}{10}\selectfont
\centering
\begin{tabular}{p{5cm}ccp{3.3cm}p{3.3cm}}
\hline
\textbf{Model} & \textbf{Mistral} & \textbf{GPT 3.5} &  \textbf{text-embedding-3-small} & \textbf{text-embedding-3-large} \\
\hline
Text only & 0.74 & 0.77 & 0.73 & 0.72 \\
+ D& 0.73 & 0.76 & 0.71 & 0.68 \\
+ D + H& 0.71 & 0.74 & 0.69 & 0.66 \\
+ H& 0.70 & 0.72 & 0.67 & 0.63 \\
+ S& 0.69 & 0.71 & 0.66 & 0.67 \\
+ D + S& 0.67 & 0.69 & 0.64 & 0.61 \\
+ H + S& - & - & - & 0.65 \\
+ PD + H + S& - & - & - & 0.62 \\
+ D + H + S& \textbf{0.65} &\textbf{ 0.68} & \textbf{0.62} & \textbf{0.58} \\
\hline
\end{tabular}
\caption{\label{model-results}
Comparison of mean absolute error across different model configurations (dev set results). Ablations that included both annotator history and survey information were only performed on the best-performing model. D refers to Annotator Demographics, H refers to other texts an annotator has rated, S refers to survey responses, PD refers to predicted demographics.
}
\end{table*}
\label{sec:collaborative_filtering}
\begin{table*}
\fontsize{9}{10}\selectfont
\centering
\begin{tabular}{p{12cm}c}
\hline
\textbf{Experiment Description} & \textbf{Individual MAE} \\
\hline
Initial training with Collaborative Filtering approach and RoBERTa & 1.12 \\
Adjusted annotation embedding dimensions from 8 to 512 & 0.89 \\
Freezing RoBERTa after pre-training on \citep{KUMAR2021DESIG} & 0.80 \\
\hline
\end{tabular}
\caption{\label{important-experiments} Significant Experiments and Their Impact on Mean Absolute Error (MAE)}
\end{table*}

\label{sec:approaches_taken}
\label{sec:appendix}
\textbf{Approaches Taken}
\begin{enumerate}
    \item Tried to cluster the annotator embeddings (PCA) – they weren’t linearly separable based on demographics
    \item Where to incorporate recommender systems
    \begin{enumerate}
        \item Classification head start – features
        \item Later layer
        \item  before appending to `features`
    \end{enumerate}
    \item Tried to train plan RoBERTa on the entire dataset using the pretrained\_multitask\_demographic dataset
    \item Different dimensions of annotator embeddings
    \begin{enumerate}
        \item Tried dim 8: little to no predictive power for annotator demographics
        \item Changed to 512
        \item Now using dim 768
    \end{enumerate}
    \item Dual RoBERTa
    \begin{enumerate}
        \item Instead of randomly instantiating an embedding layer, we tried using RoBERTa to represent annotators based on their IDs.
    \end{enumerate}
\end{enumerate}

\textbf{Text Structure}

For these predictions, the input is formatted as $h_1 \dots h_n$ \; \texttt{[SEP]} \; $s_1 \dots s_n$ \; \texttt{[SEP]} \; $d_1 \dots d_n$ \; \texttt{[SEP]} \; $w_1 \dots w_n $, where $h_1 \dots h_n$ represents the other texts reviewed and their ratings as provided by the annotator, $s_1 \dots s_n$ is a template string describing the annotator's survey information data, $d_1 \dots d_n$ is a template string containing the annotator's demographic information (e.g., ``The reader is a 55-64 year old white female who has a bachelor’s degree, is politically independent, is a parent, and thinks religion is very important. The reader is straight and cisgender''), $w_1 \dots w_n$ is the text being rated, and \texttt{[SEP]} is a separator token. We use a template string instead of categorical variables in order to best take advantage of the model's language pretraining objective (e.g., underlying associations about the experiences of different demographic groups). 

\textbf{Dataset Size}

The dataset we used to evaluate the performance of our approaches -- \citep{KUMAR2021DESIG} -- has 3 splits: train, dev, and test. The training set has 488,100 samples, the dev set has 25,000 samples, and the test set also has 25,000 samples.

\textbf{Model Information}

For the collaborative filtering approach, we used a RoBERTa model that has 355 million trainable parameters, and it took 2 GPU hours per epoch when fine-tuned on 2 NVIDIA Quadro RTX 8000 GPUs.

For the ICL approach, we used an API version of OpenAI's text-embedding-3-large model, which we don't have access to, so as to determine its size, and infrastructure requirements.

\textbf{Experimental Setup}

We observed the best performance when having 4 dense layers after the embedding was outputted, which transformed the embedding from 3072 dimensions to 1024 dimensions, then keeps it at 1024 dimensions for another 2 layers after which the last layer is then shrunk to 5 dimensions.

\textbf{Demographics Prediction Task}
\begin{table*}[]
\resizebox{\textwidth}{!}{%
\begin{tabular}{lcccccc}
\hline
\textbf{Generated Data} & \textbf{Race } & \textbf{Gender} & \textbf{Importance of Religion} & \textbf{LGBT Status} & \textbf{Education} & \textbf{Political Stance} \\ \hline
Survey Info            & 47\% & 63\% & 37\% & 38\% & 57\% & 48\% \\
Survey Info + Text     & 43\% & 60\% & 33\% & 34\% & 52\% & 44\% \\ 
Majority Class         & 9\%  & 52\% & 31\% & 81\% & 52\% & 40\% \\ \hline
\end{tabular}%
}
\vspace{-7px}
\caption{\label{table:generated-data-results}
Comparison of demographic prediction accuracy across different data configurations.
}\end{table*}
\textbf{Figure 1: Neural Collaborative Filtering}

\begin{figure}[t]
\centering
\fbox{
\begin{minipage}{0.95\columnwidth}
\textbf{\color{teal}System Prompt}

You are a model that predicts the toxicity rating of text from 0 to 4, where 0 is the least toxic and 4 is the most toxic.

\vspace{0.3cm}

\textbf{\color{olive}User Prompt}

The annotator has annotated these texts: ``This is a harmless comment'' is rated as 0, ``You're an idiot'' is rated as 3, ``I respectfully disagree'' is rated as 1 [SEP] The reader uses social media, news sites, and video sites. The reader has seen toxic comments, has been personally targeted by toxic comments, thinks technology has a positive impact on people's lives, and thinks toxic comments are a serious problem. [SEP] The reader is a 25-34 year old Asian female who has a Bachelor's degree, is politically liberal, is not a parent, and thinks religion is not important. [SEP] Annotate this text: ``Why don't you go jump off a cliff?''
\end{minipage}
}
\caption{Sample prompt for toxicity prediction model. The system prompt (in teal) defines the model's role. The user prompt (in olive) provides historical annotations, survey results, demographic information, and the text to be rated.}
\label{fig:toxicity_prompt}
\end{figure}

\end{document}